\documentclass[runningheads]{llncs}

\usepackage[table]{xcolor}
\usepackage{tikz}
\usetikzlibrary{shapes, arrows, positioning}
\usepackage{longtable}
\usepackage{tabularx}
\usepackage{algorithm}
\usepackage{booktabs}
\usepackage{algpseudocode}
\usepackage{enumitem}
\usepackage{comment}
\usepackage{graphicx}
\usepackage{subcaption}
\usepackage{dsfont}
\usepackage{multirow}

\usepackage{hyperref}
\hypersetup{colorlinks,allcolors=blue}

\definecolor{lightpurple}{RGB}{230, 220, 250}
\definecolor{purple}{RGB}{150, 120, 180}

\usepackage[square,numbers,sort]{natbib}
\makeatletter
\renewcommand\bibsection%
{\section*{\refname
    \@mkboth{\MakeUppercase{\refname}}{\MakeUppercase{\refname}}}
}
\renewcommand\@biblabel[1]{#1.}

\makeatother
\usepackage{amsmath,amssymb}
\renewcommand{\vec}[1]{\ensuremath{\mathbf{#1}}}
\usepackage{mathtools}
\usepackage{booktabs}

\usepackage[
inline,
]{newrevisor}

\newrevisor{manuel}{violet!75}
\hiderevisor{manuel}

\newrevisor{josh}{cyan}
\hiderevisor{josh}

\newrevisor{matt}{teal}
\hiderevisor{matt}

\begin{document}

\title{An adaptive approach to Bayesian Optimization with switching costs}

\author{Stefan Pricopie\inst{1} \and
Richard Allmendinger\inst{1} \and
Manuel López-Ibáñez\inst{1}\and
Clyde Fare\inst{2} \and Matt Benatan\inst{3} \and 
Joshua Knowles\inst{1}}
\authorrunning{S. Pricopie et al.}

\institute{Alliance Manchester Business School, University of Manchester, United Kingdom\\ \and
IBM, United Kingdom\\ \and
Sonos\\}
\maketitle


\begin{abstract}
% While the evaluation cost of a solution may be financial or computational, traditionally it is assumed to be homogeneous and unaffected by the history of previous evaluations.
We investigate modifications to Bayesian Optimization for a resource-constrained setting of sequential experimental design where changes to certain design variables of the search space incur a switching cost.
This models the scenario where there is a trade-off between evaluating more while maintaining the same setup, or switching and restricting the number of possible evaluations due to the incurred cost.  
We adapt two process-constrained batch algorithms to this sequential problem formulation, and propose two new methods --- one cost-aware and one cost-ignorant.
We validate and compare the algorithms using a set of 7 scalable test functions in different dimensionalities and switching-cost settings for 30 total configurations.
% We find that switching setups every 2 or 3 evaluations performs well in most use cases, however, such resource planning strategies are restrictive and do not adapt to the algorithm's performance.
Our proposed cost-aware hyperparameter-free algorithm yields comparable results to tuned process-constrained algorithms in all settings we considered, suggesting some degree of robustness to varying landscape features and cost trade-offs. This method starts to outperform the other algorithms with increasing switching-cost. 
Our work broadens out from other recent Bayesian Optimization studies in resource-constrained settings that consider a batch setting only. 
While the contributions of this work are relevant to the general class of resource-constrained problems, they are particularly relevant to problems where adaptability to varying resource availability is of high importance.

\end{abstract}

\section{Introduction}

In expensive black-box optimization problems, some decision variables may be more costly to change than others, leading to different evaluations having different costs depending on which decision variables are changed. 
For example, in the automotive or electronics manufacturing industries, whenever there is a need for a new product, a retooling of the production line needs to be implemented~\citep{teunter_multi-product_2008}.
% Each retooling change requires more time and resources on top of the expensive evaluation.
A line switch can delay production time due to an idle period and lead to additional setup costs from any specialized labor employed for instrument recalibration, equipment cleaning, or replacement of consumables.
In such settings frequent action changes are prohibitive as the cumulative switching cost between setups will rapidly increase, limiting the number of evaluations possible. A switching cost is encountered every time there is a change in the setup.

In several problems, we see that the history of the optimization process influences the cost and resources available. 
While in Bayesian Optimization (BO) it is typical to allow decision variables to be changed freely between iterations this scenario does not fully capture the logistical constraints of real-world applications: the cost of changing any decision variable is not always the same.
% many involving physical limitations such as the unavailability of goods or high purchasing costs.
% Traditionally changes in the control parameters are considered free, however, this is not always the case in physical experiments.
It is expected that for any changes made throughout the optimization process, particularly in dynamic or time-dependent problems, some penalty is incurred, whether that is an increase in time, cost, resource use, or any other variables involved.
Examples include heat-treatment scheduling~\citep{vellanki_process-constrained_2017}, robot design~\citep{liao_data-efficient_2019}, molecular discovery~\citep{pricopie_expensive_2022} or path-based problems in environmental sciences~\citep{folch_snake_2022}. 
In such problems, the history of the most recent evaluation(s) makes certain neighboring points cheaper to evaluate than others, thus making exploitation cheap and exploration expensive.

Similar scenarios are encountered in problems around chemical processing plants~\citep{behr_new_2004} and biopharmaceutical production~\citep{eberle_improving_2014} when there is a change in formula, in food processing facilities changing between recipes~\citep{kopanos_resource-constrained_2011}, or during large construction projects with multiple phases of completion~\citep{aslam_design_2019}. 
The same issues are also encountered in supply chain logistics when there is a need for a reconfiguration of the distribution network, a switch of the transportation mode, or an inventory uncertainty driven by supply levels~\citep{dev_hybrid_2016}. 

Our work was motivated by such restrictions in experiments involving physical resources that are usually limited and often influence the optimization cost and the overall process.

In many optimization problems, we can distinguish two types of costs: 

\begin{enumerate}[label=(\roman*)]
    \item an evaluation cost, which can be fixed or variable depending on the problem formulation; and
    \item a setup cost, which involves the preparation before the evaluation.
\end{enumerate}
While there has been substantial research on homogeneous and heterogeneous evaluation costs, there has been little research on setup costs and bottlenecks arising from performing physical evaluations. The main decision then becomes when and how often a switch to a cost-associated decision variable should happen given the optimization time, the number of resources used, and the improvement in the solution quality achieved.
% \MANUEL{You introduce these definitions but never say how they apply to previous works and how they apply to our work.}\MANUEL{If I understand correctly, in this paper, there is a setup cost only for changing particular which variables are changed relative to the previous evaluation.}
% A setup cost is encountered when the optimization process assumes a requirement for new materials or setups, for example, a new reconfiguration in a manufacturing production line. 
% While the two costs are to be found to have a significant impact on the optimization, little research has been conducted to understand their behavior.  

An example of such resource-constrained BO was proposed by~\citet{vellanki_process-constrained_2017} in the form of process-constrained batch optimization. 
Their experimental setup imposes constraints on the batch construction by only allowing batches that share the same values for some decision variables.  
In this paper, we are extending their formulation to a generalized form by transforming the batches into a sequential problem. Since we are looking at sequential rather than batch settings, the resource constraints are affected by previous evaluations throughout the search space, which is not the case in the batch setting where the constraint affects only the respective batch.
We also soften their constraint by allowing switches with a penalty associated with it. The formulation by \citet{vellanki_process-constrained_2017} can then be seen as a particular case of our problem where the switching cost is $+\infty$, even though a cost is not specifically modeled in their case.

% \begin{figure}[hbt!]
%     \centering
%     \includegraphics[width=0.6\columnwidth]{Figures/pfbo_0.10.pdf}
%     \caption{Probabilistic Fixing BO where the axes $x_0$ and $x_1$ represent the expensive and cheap dimensions respectively. The colour of the points relates to the sequence of evaluations between 0 and 20, and $p$ is set to 0.1. The dotted lines mark reusing the same setup (no switching) from the previous evaluation.}
%     \label{fig:pf_plot}
% \end{figure}

The overarching research question addressed in this paper is whether there are good adaptive and/or parameter-free strategies an optimizer can use to
choose between evaluating cheaply (by reusing the same setup as before) and changing the setup altogether for a more expensive evaluation. Every change to a new setup will incur a switching cost, and we wish to be robust to different \emph{relative}\/  switching costs.

The main contributions of the paper are:
\begin{enumerate}
    \item We extend the optimization problem proposed in~\cite{vellanki_process-constrained_2017} from a batch-constrained formulation to a sequential one to allow for broader generalizability. Our formulation is resource-aware by penalising frequent changes performed in the costly dimension and considers the setup cost for more thorough optimization.
    \item We show how the optimization process is affected by both the setup and the evaluation cost. Ignoring either of the two can lead to suboptimal convergence with respect to the overall cost of optimization.
    \item We adapt the two resource-unaware algorithms proposed by~\citet{vellanki_process-constrained_2017} for batch constraints to resource-aware sequential settings. Additionally, we propose two new algorithms for this problem, an Expected Improvement per Unit Cost (EIPU) and a Probabilistic re-use BO. 
    \item We show that previously proposed solutions require restricting the optimization to several evaluations per setup. The choice of this number is nontrivial and depends on the setup cost and the dimensionality of the problem.
\end{enumerate}

The rest of this work is organized as follows. In Section~\ref{sec:literature} we review the related work. The problem setup is formally introduced in Section~\ref{sec:problem}. In Section~\ref{sec:methodology} we describe the algorithms used, with the experimental setup given in Section~\ref{sec:experiments} together with the results of experiments. Finally, we discuss any limitations and directions for future research in Section~\ref{sec:conclusion}.

\section{Related Work}\label{sec:literature}
%The work done in this paper falls closest within
Our research builds on the academic literature of process-constraint BO, and more specifically studies that consider cost, for example, per evaluation or setup, as part of their objective function. 
% BO is a popular framework used predominantly in black-box expensive optimization \citep{frazier_tutorial_2018} and has been applied to a range of applications such as hyperparameter tuning in machine learning \citep{snoek_practical_2012}, drug discovery \citep{griffiths_constrained_2019, griffiths_constrained_2020}, or materials design \citep{zhang_bayesian_2020}. 
While the literature around BO is extensive~\citep{frazier_bayesian_2016, griffiths_constrained_2020, zhang_bayesian_2020}, there is not an equal focus on those BO applications that use a non-fixed cost. 
The cost per evaluation has been a relatively more popular research problem, both in the multi-fidelity~\citep{swersky_multi-task_2013, kandasamy_gaussian_2016, song_general_2019} and single-fidelity space~\citep{lin_bayesian_2021,lee_nonmyopic_2021,astudillo_multi-step_2021}. 
With regard to incurring a switching cost for new setups, we have identified several papers that are directly related to our research~\citep{snoek_practical_2012, vellanki_process-constrained_2017, liao_data-efficient_2019, lin_bayesian_2021, folch_snake_2022}. 

\citet{snoek_practical_2012}'s seminal paper emphasized BO's potential on hyperparameter tuning in neural networks while being the first to bring to the attention the problem of cost in optimization. 
Whenever there is an expensive optimization problem, the assumption is usually that this cost is fixed. 
Nevertheless, the cost per evaluation in neural networks is proportional to the batch size or other pre-defined parameters. 
For example, choosing an evaluation over a batch size of 10 would be considerably cheaper in terms of time than with a batch of 100. 
Hence, \citet{snoek_practical_2012} introduced the (EIPU) \citep{lee_cost-aware_2020}, which nonetheless assumes that each evaluation requires a new setup and does not explicitly model it in the problem.

% \citet{vellanki_process-constrained_2017} is setting the problem of evaluation costs in the context of metallurgy where you might want to optimize both the amount of material cast and the temperature. 
% In a manufacturing setting, you may be forced to process some heat treatments on materials at certain temperatures. 
% Once the furnace is set at a certain temperature, you can only allocate batches that require the setting of the machine in terms of temperature. 
% The main idea is that because a process is restrictive and some parameters are harder to change than others, to save time they perform evaluations in batches. 
% The authors propose two approaches to dealing with the constraints, namely a basic and a nested process-constrained BO (pcBO-basic and pcBO-nested). 

% Our paper differs from \citet{vellanki_process-constrained_2017} in that it looks at sequential rather than batch optimization. \citet{vellanki_process-constrained_2017}'s objective function is written in terms of the number of batches used. They are trying to find the algorithm that arrives at the optimal solution in the fewest number of batches. However, this assumes that there is a setup cost, but it does not account for the evaluation cost. The size of the batch becomes then irrelevant in their approach. 

\citet{liao_data-efficient_2019} extends~\citet{vellanki_process-constrained_2017}'s batch construction, where batches need not be constrained to a single setup and can be parallelised. 
In this case multiple batches are running in parallel, which constrains the optimization process based on the setup cost of $n$ number of machines (robots in this case) for a while. 
Since morphologies are expensive and take a lot of time to produce, the goal of the paper is to evaluate multiple morphologies (microrobots) in parallel and optimise their unconstrained parameters (here controllers) while minimizing cost (how many morphologies are produced). 
Similar to~\citet{vellanki_process-constrained_2017}, their algorithms are optimised against the number of new setups and do not account for evaluation costs, meaning that the only difference between a higher batch and a lower batch is a computational cost.

To the best of our knowledge,~\citet{lin_bayesian_2021} are the first to introduce the idea of switching costs in the context of BO in a continuous space. 
They partition the search space into disjoint tiles. Their cost is then measured when the optimization process jumps from one tile to another. 
Raising similar issues to this problem is also the work presented by~\citet{folch_snake_2022}. 
While both papers discuss the scenario where the cost of your optimization arises in between evaluations,~\citet{folch_snake_2022} considers the case where the cost is the distance between evaluations, thus making (proximal) exploitation cheaper and (distal) exploration more expensive. 

% \citet{calandriello_scaling_2022} is looking at the context of scaling up a Gaussian Process (GP) which is one of the main challenges of GPs. 
% Their objective is to train the GP with a large amount of data points, while also allowing for repeated evaluations. 
% This is different from optimization, where there are few evaluation points. 
% Therefore, in optimization, the budgeting strategy needs to be adaptable to the resulting evaluations. 
% They suggest a problem where there is an experimental setup. 
% Instead of restarting the setup at each evaluation, it is preferable to do multiple evaluations over one noisy evaluation point. 
% This scenario implies that the problem has noise leading to different results, which is commonly met in real-world applications. 
% Their work differs from our approach in two main ways. 
% Firstly, the scope of the paper is not to minimise cost but rather to create a scaled-up GP. 
% Secondly, since scalability is their main objective, they are not interested in the results obtained when evaluating in the same place. 
% We care about previous results to decide whether it is worth continuing to evaluate in the same place or whether a new setup is required.

Our problem differs from previous research in a few distinct ways. Unlike the majority of previous literature, we are considering a sequential setting rather than a batch one. This approach was preferred because it makes clearer a difficulty that is more hidden in the batch setting. Optimising in a batch scenario you are constrained to evaluating multiple points and deciding a priori where to evaluate. In a sequential setting, the constraint is not enforced so that the number of available optimization points is sufficiently high to create a trade-off problem. We further note that with additional work the sequential setting can be reduced to a batch one should further constraints on how evaluations are grouped be imposed. Moreover, another main limitation of the batch approach is that it becomes difficult to identify which is the optimal batch. Once a setup is established, there is no quantifiable way to decide how many evaluations you should perform within the same batch before changing.

Our approach also differs from previous papers by adding the switching cost element. Previous research considers the problem as a hard constraint, meaning that once you have a point of evaluation, this cannot be changed for the time being. We rather look at the problem assuming a soft constraint, meaning that you can evaluate in any other place within the search space, but this will incur an extra cost.

% \begin{figure*}[ht]
%     \centering
%     \begin{subfigure}{0.32\textwidth}
%         \includegraphics[width=\linewidth]{Figures/lin_2021.jpg}
%         \caption{\citet{lin_bayesian_2021}}
%         \label{fig:sub1}
%     \end{subfigure}
%     \hfill
    % \begin{subfigure}{0.32\textwidth}
    %     \includegraphics[width=\linewidth]{Figures/vellanki.jpg}
    %     \caption{\citet{vellanki_process-constrained_2017}}
    %     \label{fig:sub2}
    % \end{subfigure}
    % \hfill
%     \begin{subfigure}{0.32\textwidth}
%         \includegraphics[width=\linewidth]{Figures/samaniego.jpg}
%         \caption{\citet{samaniego_bayesian_2021},\citet{folch_snake_2022}}
%         \label{fig:sub3}
%     \end{subfigure}
%     \caption{To be replaced with original figures}
%     \label{fig:three-images}
% \end{figure*}

\section{Problem Setup}\label{sec:problem}
This paper considers expensive optimization problems on \(f\colon \mathcal{X} \to \mathbb{R}\) where the domain
$\mathcal{X} = \mathcal{X}_\text{cheap} \oplus \mathcal{X}_\text{costly}$ is the direct sum of the vector spaces $\mathcal{X}_\text{cheap}$ and $\mathcal{X}_\text{costly}$ with different computational costs.
The dimensionality of $\mathcal{X}$ is then the sum of its subspaces, $d=d_{\text {cheap}} + d_{\text {costly}}$, where $d_{\text {cheap}}$ and $d_{\text {costly}}$ are the dimensions of the two vector spaces respectively. The cost function \(c\colon \mathcal{X} \times \mathcal{X} \to \mathbb{R} \) is given by:
\begin{equation}
    c(\mathbf{x}^t,\mathbf{x}^{t-1}) = \begin{cases} 
    c_\text{switch} & \text{if } \mathbf{x}_\text{costly}^t \neq \mathbf{x}_\text{costly}^{t-1}, \\
    1 & \text{otherwise}.
    \end{cases}
    \label{eq:cost}
\end{equation}

\noindent where $\mathbf{x}^t=\left[\begin{array}{ll}
\mathbf{x}_{\text {cheap }}^t & \mathbf{x}_{\text {costly }}^t
\end{array}\right]$ is the $t^{\text{th}}$ function evaluation and $c_\text{switch} \in [1, \infty)$.
We refer to a setup change or a switch when $\mathbf{x}_\text{costly}^t \neq \mathbf{x}_\text{costly}^{t-1}$. For example, when $c_\text{switch} = 5$,  it is $5$ times more expensive to switch than to reuse the same setup.
When $c_\text{switch} = 1$,  it is equivalent to the traditional optimization problem.
Intuitively, $c_\text{switch} - 1$ is the penalty for changing the setup for a new experiment while $c_\text{switch}$ is the penalty multiplier. 

\begin{figure}
    \centering%
    \includegraphics[width=0.9\linewidth]{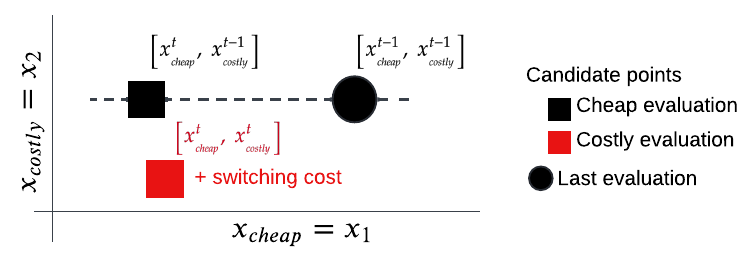}
    \caption{Example of a $d=2$ problem with $d_\text{costly} = 1$. The x-axis and y-axis represent the cheap and costly dimensions, respectively. The dotted line is the set of evaluations that use the current setup. The squares are the candidate points and the dot is the last evaluation.}
\label{fig:switching_cost}
\end{figure}

% In this paper, we present the problem of BO with switching costs. In a switching cost scenario, we have certain dimensions which give us a setup cost. These dimensions are generally expensive to evaluate and hence whenever we make a change a pre-defined cost associated with it will be incurred.\MANUEL{The previous sentences sound like an introduction but you have already explained that you are considering BO with switching costs so no need to explain it again. } 
In the example shown in Figure~\ref{fig:switching_cost} the black dot is the last evaluation and the square is the best next point to evaluate, which is on the line, meaning that the same setup applies and there is not a new cost associated with the move. However, evaluating at the red square, which is the best next point to evaluate, is within a different setup, hence the function evaluation will have a new setup cost.

\begin{figure}[t!]
    \centering%
    \includegraphics[width=0.6\columnwidth]{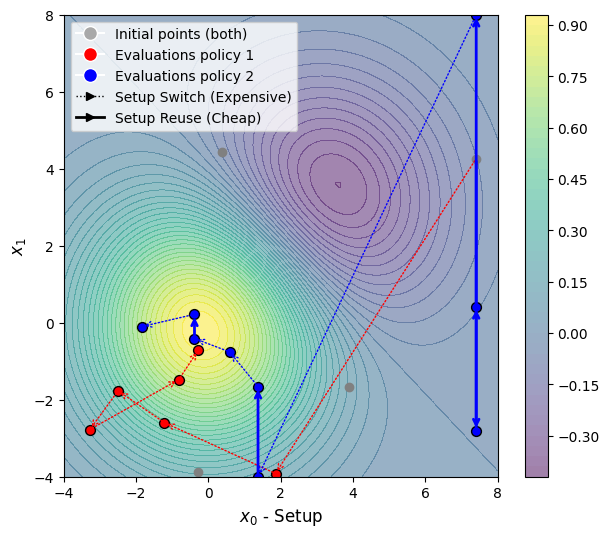}
    \caption{Maximization problem on synthetic function with axes $x_0$ and $x_1$ respresenting the expensive and cheap dimensions in the search space respectively. The grey dots represent the initial starting points, which are randomly selected. The red and blue lines then denote two evaluation strategies. The red line represents a strategy where the algorithm changes the setup at each evaluation, while the strategy represented by the blue line uses the same setup for evaluations 2, 3, 5, and 8. The second evaluation policy performs 9 evaluations, while the first policy performs only 6 due to its more expensive nature. Reusing the same setup allows for more total evaluations at the expense of fewer evaluations performed across the costly dimension $x_0$.}
    \label{fig:pf_plot}
\end{figure}

In Figure~\ref{fig:pf_plot} we are using a synthetic test function to show the behavior of a solution method in the cases of setup switching costs. Depending on the next chosen point to evaluate, the algorithm can select a point which incurs an evaluation cost. For example, in evaluation policy 1, there is a switching cost associated to every move made, while in the evaluation policy 2, there are 3 moves that do not have a switching cost associated since the same setup is reused. This type of scenario then raises the question of whether the algorithm can find an optimal trade-off between the costs associated with a switch and the quality of the solution found. 

\section{Methodology}\label{sec:methodology}

We are interested in characterising the trade-off between evaluating the cheap dimensions for more evaluations and evaluating the costly dimensions for less. We investigate whether it is possible to find an optimal $p^\ast$ value (or range), where $p$ is the probability of re-using a setup and hence evaluating again on the cheap dimension without incurring a switching cost. We analyze four algorithms capable of tackling this problem, in addition to a classic BO that ignores the switching cost.

% Classical BO is cost-unaware of the problem so therefore each evaluation would perform a new switch.
% In other words, the likelihood of classical BO evaluating a point that only changes the cheap dimensions is very close to zero.
% Intuitively, if the algorithm performs switching all the time, meaning that $p = 0$, it will be suboptimal. 
% Gathering more information through a higher number of cheaper evaluations is preferred over performing a few expensive ones. 
% In other words, there is thus a trade-off between the number of evaluations you can afford given a certain budget, cost, dimensions, and the number of switches. 

The first new algorithm we propose is \textit{pReuseBO}.
This algorithm keeps the same expensive decision variables independently with probability $p \in [0, 1]$ every time step. The $p$ is set to 0.1 which on average would lead to 1 out of 10 points being evaluated on the cheap dimension, meaning that the rest 8 would incur a switching cost. Figure~\ref{fig:pf_plot} illustrates the behavior of this algorithm for 20 evaluations in a two-dimensional space.
In the given example, there are 4 points that do not incur a switching cost.

\begin{figure}[hbt!]
    \centering%
    \includegraphics[width=0.9\columnwidth]{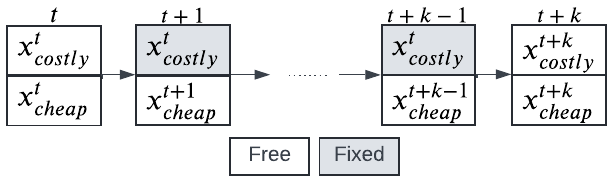} 
    \caption{Periodic switching of $k$. The costly decision variables are changed only every $k$ evaluations.}
    \label{fig:ps_plot}
\end{figure}

The difference in our problem to previous research (e.g., \citep{vellanki_process-constrained_2017}) is that instead of batches, we treat the problem sequentially. Once we choose an optimal point to evaluate by looking at the entire search space, we automatically evaluate it without creating a batch with variations. Since we evaluate a single point, now we are constrained to a $k-1$ number of evaluations. $k$ is a simpler planning strategy to the above $p$ representation, where the setup is changed after every $k$ evaluations. $k$ represents the periodicity to which an event happens and is graphically presented in Figure~\ref{fig:ps_plot}. 
In other words, the algorithm keeps the same setup for $k-1$ steps out of $k$. This is different to batching because we refit the GP after each evaluation. One possible approach is to maintain the value of the costly dimensions with probability $1-p$ or change them with probability $p$ (and pay the switch cost). Another possible approach is to maintain the value of the costly dimensions for $k$ evaluations, then change them (and pay the switch cost) and perform another $k$ evaluations only changing the cheap dimensions. We would expect that setting $k = \frac{1}{1-p}$ will produce a similar behavior on average.

We are also adapting to our problem two algorithms proposed by~\citet{vellanki_process-constrained_2017}. The first algorithm, PSBO, starts with a pre-defined number of batches. They use a GP with an upper confidence bound (UCB) as the acquisition function to search the entire search space until the optimal point is found. For each chosen point to evaluate, it is then decided which are the constrained parameters ($\mathbf{x}^t_\text{costly}$) and then they are fixed. To construct the batch, for $k$ evaluations \citet{vellanki_process-constrained_2017} create a set of points that differ in the cheap decision variables. The process is then repeated until the desired number of evaluations is exhausted. Once a point is chosen with the UCB, the acquisition function is then changed to a posterior standard deviation (PSD). We substitute the two acquisition functions with EI because both UCB and PSD (also known as Pure Exploration) can be too exploratory \citep{de_ath_greed_2019}.

The second algorithm adapted from \citet{vellanki_process-constrained_2017} is shown in Algorithm~\ref{alg:psbo_nested}. The initialization of this algorithm is presented in Algorithm~\ref{alg:nested_initiation}, where $\alpha$ is the acquisition function. The algorithm has a nested approach, by solving two problems independently. 
For each batch, \citet{vellanki_process-constrained_2017} perform an outer stage evaluation to find all the constrained points, and an inner optimization for the unconstrained points \citep{vellanki_process-constrained_2017}. 
This approach helps by dividing the search space since you have two BO algorithms running at the same time. 
The rest of the algorithm is the same as the first algorithm. 
Keeping the GP-UCB followed by a PSD was preferred in both algorithms given that with the GP-UCB they are looking to mostly exploit, while with the PSD it is mainly an exploration of the search space. 
However, it was not empirically evaluated in the paper that this approach yields better results.

\begin{algorithm}[t!]
\caption{Initialization for Periodic Switching BO Nested}
    \label{alg:nested_initiation}
\begin{algorithmic}[1]
\State \textbf{Input:} $f$, $\mathcal{X} = \mathcal{X}_\text{cheap} \oplus \mathcal{X}_\text{costly}$, \Comment{Initial ``costly'' points $n$, periodicity length $k$}

\State Select unique ``costly'' points $\{\mathbf{x}_\text{costly}^{1}, \dots, \mathbf{x}_\text{costly}^{n}\}$, each to be used $k$ times
\State Initialize a set of $n \times k$ unique ``cheap'' points $\{\mathbf{x}_{\text{cheap}}^t\}_{t=1}^{n \times k}$
\State Define the initial points $\mathbf{x}^t = \begin{bmatrix} \mathbf{x}_{\text{cheap}}^t & \mathbf{x}_\text{costly}^{\lceil t/k \rceil} \end{bmatrix}$ for $t = 1, \dots, n \times k$
\State Evaluate $f$ at each $\mathbf{x}^t$ to get $y^t = f(\mathbf{x}^t)$ for $t = 1, \dots, n \times k$
\State Form the complete dataset $D = \{(\mathbf{x}^t, y^t)\}_{t=1}^{n \times k}$

\State \textbf{Return} $D$
\end{algorithmic}
\end{algorithm}

\begin{algorithm}[t!]
\caption{Periodic Switching BO Nested}
    \label{alg:psbo_nested}
\begin{algorithmic}[1]
\State \textbf{Input:} Initial dataset \(D\), \( t = |D| \) \Comment{From Algorithm \ref{alg:nested_initiation}}

\While{stopping criteria not met}
    \If{\(t \mod k = 0\)}
        \State \(D_{\text{costly}} = \left\{ \left( \mathbf{x}_{\text{costly}}^i, \max \left\{y \mid \left([\mathbf{x}_{\text{cheap}}, \mathbf{x}_{\text{costly}}^i], y\right) \in D\right\} \right) \mid i \in \overline{1,t/k}\right\}\)
        \State Train \(GP_{\text{costly}}\) using the dataset \(D_{\text{costly}}\)
        \State \( \mathbf{x}_\text{costly} = \arg\max_{x} \alpha_\text{costly}(x ; D_\text{costly})\) \Comment{Choose new costly parameter}
    \EndIf

    \State Train \(GP\) using the dataset \(D\)
    \State \( \mathbf{x}_\text{cheap}^{t+1} = \arg\max_{x} \alpha(x ; D \mid \mathbf{x}_\text{costly} )\) \Comment{Condition on last costly parameter}
    \State Evaluate \(y_{t+1} = f([\mathbf{x}_\text{cheap}^{t+1}, \mathbf{x}_\text{costly}])\)
    \State Update \(D\) with \(([\mathbf{x}_\text{cheap}^{t+1}, \mathbf{x}_\text{costly}], y_{t+1})\)
    \State \(t = t + 1\) \Comment{Increment iteration counter}
\EndWhile
\end{algorithmic}
\end{algorithm}

We also propose an adaptation of the EI per unit cost (EIPU) to the pathological case where the cost function is discontinuous. As we are incorporating cost, and since we want to compare the penalty incurred while evaluating outside of the initial setup, we want to adapt the EIPU. 
Unlike previous methods used for this problem type, this is the first algorithm that is cost-aware.
% We are then moving the methodological focus towards being considerate of the optimization costs resulting from different moves. 
In a traditional sense, EI is a continuous function, which matches the general formulation of cost in the literature as continuous. However, in our case, cost is a discontinuous function (Eq.~\ref{eq:cost}). 
To address this we perform two acquisition optimizations within the same iteration. This will produce two candidate points for the next evaluation: (i) a candidate point that changes a costly dimension and hence incurs a switching cost and (ii) a candidate point that only changes cheap dimensions and hence without having a switching cost.
The algorithm will then discount the EI of the two points by their respective cost and the best point is chosen.
When we discount the cost we are considering the cost cooling strategy proposed in \citet{lee_cost-aware_2020}. The cost cooling strategy lessens the impact of the cost model with the advance of iterations. $\text{EI-cool}(\vec{x}) \coloneqq \frac{EI(\vec{x})}{c(\vec{x})^{\gamma}}$, where $\gamma$ is the decay parameter where $\gamma = (B - B_t) / B$, $t$ is the iteration number, $B$ is the total budget, and $B_t$ is the budget used up to time step $t$.

\section{Experiments}\label{sec:experiments}
We run two sets of experiments: (i) finding the best $p$ value across the different dimensions and switching costs using \textit{pReuseBO} (Subsection~\ref{subsec:p*}), and (ii) comparing the four algorithms' performance (Subsection~\ref{subsec:alg_performance}). The experimental setup corresponding only to a particular experiment will be explained subsequently in each respective subsection. 

\subsection{Experimental Setup}
Both sets of experiments are run on scalable test problems, shown in Table~\ref{table:test_functions}, of dimension $d = \{2, 3, 4\}$.
We choose to show the performance of the algorithms over the above test functions because the dimensionality of the problem has a direct impact on the algorithm behavior and the optimal hyperparameters. 
This argument will be empirically validated in the following sections. Moreover, while many synthetic scalable functions are symmetrical, we have included nonsymmetrical functions for robustness as symmetry is unlikely in a real-world problem. 

\begin{table}[hbt!]
\caption{Test functions considered in this study and their settings. For each function we considered the dimensions $d \in \{ 2, 3, 4\}$. For Ackley, Griwewank, and Salomon, their original domain is cropped as shown in the table, such that the optimal point was not in the centre of the domain.}
\label{table:test_functions}
\centering
\begin{tabular}{c c c}
\toprule
Functions & Symmetrical & Domain per dimension $d$ \\ 
\midrule
Ackley & Yes & $[-15, 30]$ \\ 
Griewank & Yes &  $[-300, 600]$ \\
Levy & No &  $[-10, 10]$ \\
Michalewicz & No &  $[0, \pi]$\\
Rosenbrock & No &  $[-5, 10]$ \\
Salomon & Yes & $[-50, 100]$ \\ 
Schwefel & Yes &  $[-500, 500]$\\ 
\bottomrule
\end{tabular}
\end{table}

The following configurations are considered:

\begin{description}
    \item \textbf{2D:} 1 cheap \& 1 costly;
    \item \textbf{3D:} 1 cheap \& 2 costly \textit{OR} 2 cheap \& 1 costly;
    \item \textbf{4D:} 1 cheap \& 3 costly \textit{OR} 2 cheap \& 2 costly \textit{OR} 3 cheap \& 1 costly;
\end{description}

The costly dimensions are randomly selected across 20 independent runs for robustness until the termination criteria is met. 
The values of $p$ that we chose to test are between $[0,1]$ inclusive with a step of 0.05 leading to 21 possible values.

The value of the switching cost is reflective of how expensive an evaluation becomes when switching the setup relative to the existing one at the moment of evaluation. 
For example, a switching cost of 2 means that changing the setup would make the next evaluation twice as expensive as it would be by maintaining the same setup. 
In this paper, we set the possible values for the switching cost at $\{2, 4, 8, 16, 32\}$. The first set of experiments does not use the $32$ switching cost to maintain the set of 5 switching cost across. The first set of experiments is analysing the traditional switching cost of $1$ which becomes irrelevant in the algorithms performance comparison. 

Our budget is referenced in terms of cost because we want to be able to benchmark cost-efficient algorithms with classical BO.
All experiments have a low budget of $N = 10 \times d$ switches, where $d$ is the dimensionality of the number of evaluations to perform, once the initial random sample has been set. 
Therefore, algorithms can before between $N$ evaluations, when switching at each iteration which is the case for traditional BO, and $N \times c_\text{switch}$ evaluations when no switching is performed and the algorithm reuses only the last setup from the initialization, when $p=1$ for \textit{pReuseBO}.

All the BO algorithms ran in this paper use Gaussian Processes surrogates with Matérn 5/2 kernel and Automatic Relevance Determination (ARD) as recommended by \citet{snoek_practical_2012}. We optimise the EI acquisition function with L-BFGS-B with 10 restarts and 2048 raw samples. For EIPU we optimise the acquisition function twice to get the two candidate points.

To evaluate the performance of the algorithms, we use the GAP measure, which is defined as $\text{GAP} = (y_i - y_0)/(y^\ast - y_0)$, assuming maximization of the objective function $f$ without loss of generality, where $y_i$ is the maximum observed objective function value within a single run, $y^\ast$ is the true optimal (maximum) value of $f$, and $y_0$ is the objective function value of the initial starting point, which is the same for all algorithms solving the same problem
\citep{huang_global_2006, lam_bayesian_2016, jiang_efficient_2017, jiang_binoculars_2019}.
GAP should be maximized and can be thought of as the inverse of the normalized regret and is preferred as it allows for generalizability in algorithm performance across multiple test functions. 

\subsection{Optimal probability value}\label{subsec:p*}
\begin{figure}[hbt!]
\setkeys{Gin}{height=43mm}
    % \subfloat[]
    {\includegraphics{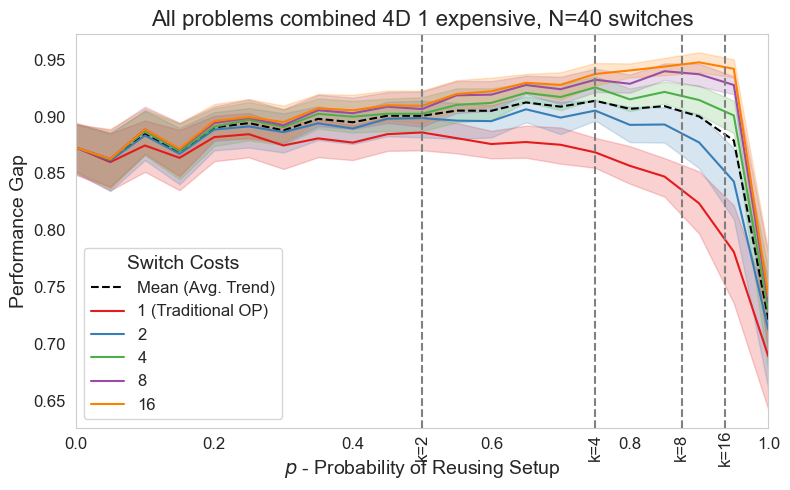}}
    \hfill% push sub images apart, so take all the line
    % % \subfloat[]
    {\includegraphics{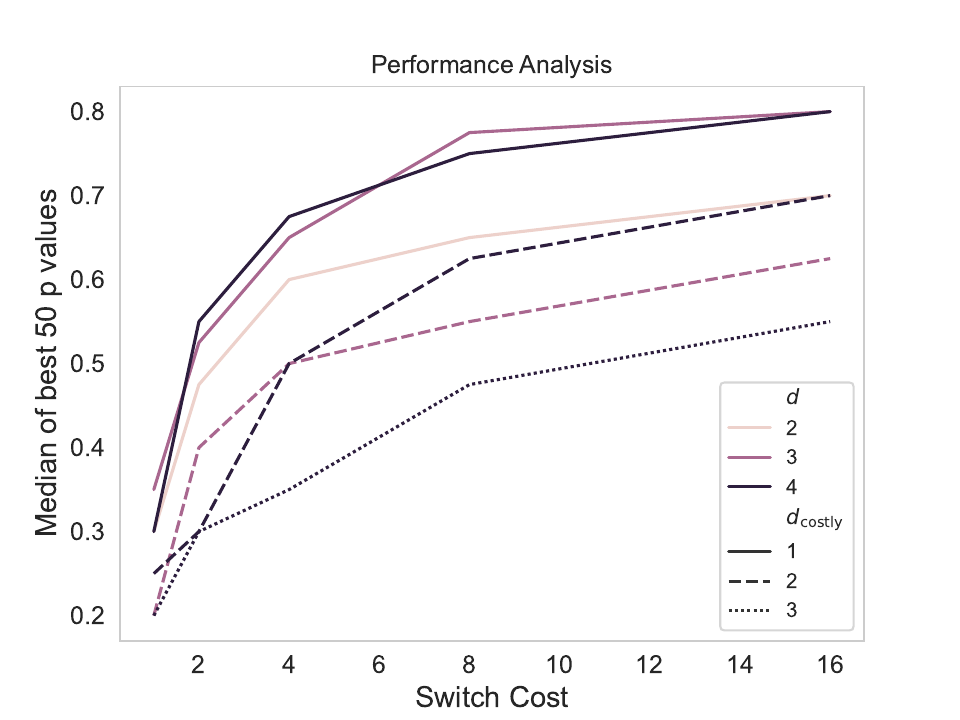}}
    \caption{
    \textbf{Left:}
    The mean GAP performance (and 95\% CI) of the method \textit{pReuseBO} plotted against 21 values of $p$, where $GAP=1$ means that the optimal solution was found, for 20 independent runs and averaged over the 7 test functions with 4 dimensions and 1 costly dimension.
    The approximately equivalent $k$ for Periodic Switching is shown on the x-axis.
    The dotted line shows the effect of $p$ averaged over all the switching costs to highlight its steady increase in performance followed by its sharp decline.
    \textbf{Right:}
    The median of the highest performing $p$ hyperparameters for each problem and costly dimensionality setting plotted against switch cost.
    $p$ is much more sensitive to tune with respect to different values of switch cost in the problems with high dimensionality and low costly dimensionality.
    }
    \label{fig:panalysis}
\end{figure}

We analyze the performance of different fixing probabilities $p$ to maximize the value of the objective function in a set budget. 
Figure~\ref{fig:panalysis} shows that an increase in the switching cost leads to an increase in $p^\ast$.
A higher switching cost means a greater trade-off between exploring the costly dimension and performing more total evaluations. 
As $p$ increases, the algorithm performance diverges. 
High $p$ values corresponding to frequent reusability perform better in high-cost settings but hinder performance in low-cost settings.

\vspace{-2em}

The optimal $p^\ast$ decreases with an increase in the costly dimensionality as shown in Figure~\ref{fig:panalysis}. 
The result is expected as the importance of exploring the costly dimension becomes higher when the number of such dimensions increases. 
The gain from performing more frequent switches in this scenario is greater, while the loss incurred from not exploring the lower number of cheap dimensions is low.

The lower the value of $p$ the closer the behavior of the algorithm is to the standard BO.
The performance of $p$ in the context of switching cost is expected to decrease quickly since there is no trade-off - one evaluation on the costly dimension is changed for one evaluation on the cheaper dimension.
This can be a reasonable trade-off for smaller values of $p$, although it is unlikely to produce a significant effect on performance.
However, with increasing $p$, there will be a steady decrease in returns and the importance of exploring the costly dimension becomes higher. 
On the other hand, an increase in the switching cost results in a higher trade-off for exchanging one expensive evaluation for multiple cheap ones.

% In the case of symmetrical functions, the cheap and costly dimensions are the same. 
% If the dimensions are the same in number, it means that with each switch that we do not perform (i.e., with each fixed feature that we keep) there is little value in evaluating the costly dimension. 
% By choosing candidate points in the cheap dimension, it will lead to the same results as evaluating in the expensive one, while allowing for the same budget to be used for a higher number of evaluations.
% In this case, a $p$ almost close to 1 can perform a very high number of evaluations at no extra cost. 

% It can also be inferred
% that the optimal value for $p$, where $p^\ast$ is the optimal probability of reusing the same setup, will increase with the budget.
% Generally speaking, the higher the budget, the higher the gains from exploiting more on the cheap dimensions.
% You can afford to perform less switches than you would in a setting where you have fewer number of evaluations available.
% With each evaluation that you perform in the low-budget setting, the gains from exploring the costly dimension are higher. 

\subsection{Algorithms performance}\label{subsec:alg_performance}

The algorithms performance is shown in Table~\ref{tab:comparison_over_best}. The experiments shown in this table are from the 4D-1 setting. We have chosen this configuration based on the results obtained over the $p$ analysis (see Figure~\ref{fig:panalysis}). These have shown that the optimal choice of reusing or switching a setup is most sensitive in high-dimensional, low expensive dimensionality settings. 

We are most interested in comparing the hyperparameter-free EI per unit cost with multiple instances of the other algorithms. Whenever there is a reference to ``best'' of an algorithm, we mean the manually tuned configuration of the algorithm that led to the best performance in terms of GAP. The findings show that despite using tuned versions of the other algorithms, these do not outperform the results obtained by the EIPU in high switching costs scenarios. In low cost setttings EIPU behaves close to BO and is outperformed by \textit{pReuseBO} and PSBO. 

In Table~\ref{tab:comparison_over_best} we highlighted the first 2 algorithms with the highest GAP achieved. In the case of switching cost 2, EIPU is outperformed for all test functions by BO Random and PSBO.
However, its performance increase in comparison to the other algorithms is evident starting with a switching cost of 4.
The EIPU is outperforming the other algorithms starting with a switching cost of 8 for the Schwefel test function. Nonetheless, for a switch cost of 16, it performs best in 4 out of the 7 test function and second best in 1 of them. Similarly, for the highest switching cost of 32, it performs best in 4 out of 7 cases and second best in 2 of them. 

Since \textit{pReuseBO} and PSBO are manually tuned, for matter of reference, we have also compared the results obtained from their untuned equivalent. Without configuring the hyperparameters of the other two methods, EIPU outperformed in almost all cases with a switching cost higher than 2. Generally, as switching cost increases, EIPU performs better as it is adaptable, exploiting good setups more. Compared to the other algorithms, the cost-aware EI switches a setup only when the current one doesn't yield large enough improvements. The other algorithms switch following a predetermined sequence by $p$ or $k$.

% We can also see that the EI per unit cost is less affected by an increasing switching cost. As expected, the behavior of the algorithms is more similar in lower switching cost settings. For some of the tested problems, the EI per unit cost outperforms the other algorithms. 

\begin{table}[hb!]
  \centering%
  \caption{Algorithm comparison for the 7 test functions, over the 5 algorithms of interest. (Best) \textit{pReuseBO}, (Best) PSBO, and (Best) PSBO Nested are the configurations who yielded the best performance in terms of GAP over 5 tested configurations for hyperparameter tuning. The algorithm with the highest GAP is highlighted in \textit{purple} and the second best is highlighted in \textit{light purple}.}
\label{tab:comparison_over_best}
\begin{tabularx}{0.88\textwidth}{ccccccc}
\toprule
\multicolumn{1}{c}{Switch} & \multirow{2}{*}{Problem} & \multirow{2}{*}{BO} & \multicolumn{1}{c}{\textit{pReuseBO}} & \multirow{2}{*}{EIPU} & \multicolumn{1}{c}{PSBO} & \multicolumn{1}{c}{PSBO Nested} \\
             \multicolumn{1}{c}{Cost}                &                          &                    & \multicolumn{1}{c}{(Best)}   &                      & \multicolumn{1}{c}{(Best)} & \multicolumn{1}{c}{(Best)} \\
\midrule

2 & ackley & 0.886344 & \cellcolor{purple}0.920431 & \cellcolor{lightpurple}0.903437 & 0.877309 & 0.760876 \\
2 & griewank & 0.991388 & \cellcolor{lightpurple}0.992109 & 0.991953 & \cellcolor{purple}0.992834 & 0.976825 \\
2 & levy & 0.993641 & 0.994666 & \cellcolor{lightpurple}0.995413 & \cellcolor{purple}0.996521 & 0.989614 \\
2 & michalewicz & 0.855549 & \cellcolor{purple}0.921913 & 0.893882 & \cellcolor{lightpurple}0.904125 & 0.741281 \\
2 & rosenbrock & 0.999413 & \cellcolor{purple}0.999773 & 0.999623 & \cellcolor{lightpurple}0.999728 & 0.999205 \\
2 & salomon & 0.868022 & \cellcolor{lightpurple}0.894309 & 0.869581 & \cellcolor{purple}0.898732 & 0.837132 \\
2 & schwefel & 0.714171 & \cellcolor{lightpurple}0.788107 & \cellcolor{purple}0.789711 & 0.784081 & 0.704648 \\

\midrule
4 & ackley & 0.886344 & \cellcolor{purple}0.928764 & \cellcolor{lightpurple}0.906928 & 0.899564 & 0.773149 \\
4 & griewank & 0.991388 & \cellcolor{lightpurple}0.992437 & 0.992114 & \cellcolor{purple}0.993251 & 0.980942 \\
4 & levy & 0.993641 & 0.995714 & \cellcolor{purple}0.996914 & \cellcolor{lightpurple}0.996546 & 0.990014 \\
4 & michalewicz & 0.855549 & \cellcolor{purple}0.950820 & \cellcolor{lightpurple}0.934351 & 0.927506 & 0.760070 \\
4 & rosenbrock & 0.999413 & \cellcolor{purple}0.999957 & \cellcolor{lightpurple}0.999950 & 0.999934 & 0.999686 \\
4 & salomon & 0.868022 & \cellcolor{purple}0.917669 & 0.906176 & \cellcolor{lightpurple}0.913210 & 0.877922 \\
4 & schwefel & 0.714171 & \cellcolor{purple}0.818559 & 0.814713 & \cellcolor{lightpurple}0.815333 & 0.726128 \\

\midrule
8 & ackley & 0.886344 & \cellcolor{purple}0.931443 & \cellcolor{lightpurple}0.929409 & 0.907566 & 0.773149 \\
8 & griewank & 0.991388 & \cellcolor{lightpurple}0.992783 & 0.991681 & \cellcolor{purple}0.994011 & 0.981282 \\
8 & levy & 0.993641 & \cellcolor{purple}0.997101 & 0.996722 & \cellcolor{lightpurple}0.996798 & 0.991616 \\
8 & michalewicz & 0.855549 & \cellcolor{lightpurple}0.960074 & \cellcolor{purple}0.983750 & 0.943899 & 0.760070 \\
8 & rosenbrock & 0.999413 & 0.999969 & \cellcolor{purple}0.999992 & \cellcolor{lightpurple}0.999973 & 0.999892 \\
8 & salomon & 0.868022 & \cellcolor{purple}0.925928 & \cellcolor{lightpurple}0.919973 & 0.916355 & 0.878876 \\
8 & schwefel & 0.714171 & 0.822432 & \cellcolor{purple}0.845480 & \cellcolor{lightpurple}0.823618 & 0.762085 \\

\midrule
16 & ackley & 0.886344 & \cellcolor{lightpurple}0.931443 & \cellcolor{purple}0.932147 & 0.907566 & 0.773149 \\
16 & griewank & 0.991388 & \cellcolor{lightpurple}0.993453 & 0.992587 & \cellcolor{purple}0.994011 & 0.981353 \\
16 & levy & 0.993641 & 0.997101 & \cellcolor{purple}0.998516 & \cellcolor{lightpurple}0.997608 & 0.993175 \\
16 & michalewicz & 0.855549 & 0.970236 & \cellcolor{purple}0.987241 & \cellcolor{lightpurple}0.977924 & 0.764227 \\
16 & rosenbrock & 0.999413 & 0.999987 & \cellcolor{purple}0.999999 & \cellcolor{lightpurple}0.999992 & 0.999922 \\
16 & salomon & 0.868022 & \cellcolor{purple}0.937830 & 0.910404 & \cellcolor{lightpurple}0.924830 & 0.905518 \\
16 & schwefel & 0.714171 & \cellcolor{lightpurple}0.840700 & \cellcolor{purple}0.867731 & 0.827956 & 0.772887 \\

\midrule
32 & ackley & 0.886344 & \cellcolor{purple}0.931443 & \cellcolor{lightpurple}0.921927 & 0.907566 & 0.773149 \\
32 & griewank & 0.991388 & \cellcolor{lightpurple}0.993453 & 0.992818 & \cellcolor{purple}0.994011 & 0.987995 \\
32 & levy & 0.993641 & 0.997158 & \cellcolor{purple}0.998989 & \cellcolor{lightpurple}0.997608 & 0.993679 \\
32 & michalewicz & 0.855549 & 0.972642 & \cellcolor{purple}0.986060 & \cellcolor{lightpurple}0.980103 & 0.764227 \\
32 & rosenbrock & 0.999413 & \cellcolor{lightpurple}0.999993 & \cellcolor{purple}0.999999 & 0.999992 & 0.999961 \\
32 & salomon & 0.868022 & \cellcolor{purple}0.945250 & \cellcolor{lightpurple}0.941213 & 0.927371 & 0.905611 \\
32 & schwefel & 0.714171 & \cellcolor{lightpurple}0.842070 & \cellcolor{purple}0.902302 & 0.827956 & 0.774452 \\

\bottomrule
\end{tabularx}
\end{table}

We observe that in a cost-aware sequential setting, PSBO does perform better than PSBO nested in any of our test functions. This is in contrast to the findings obtained by \citet{vellanki_process-constrained_2017} who have shown that PSBO nested is the better algorithm in batch settings. The difference in performance is caused by the fact that the $GP_\text{costly}$ which chooses the expensive dimension does not train well long term over the best found points for each setup. Therefore the algorithm will progress for a while but then plateau since the dimensionality reduction will prohibit the algorithm from finding the best point. Our results seem more consistent with the results from \citet{ghadimi_approximation_2018} which look at bi-level optimization methods where there is an inner and an outer loop problem. The authors highlight that a large number of iterations is required for nested algorithms to perform well.

% \begin{figure}[hbt!]
%     \centering%
%     \includegraphics[width=\linewidth]{Figures/4d1-2.png}
%     \caption{We compare the algorithm performance of EIPU with the best performing \textit{pReuseBO}, PSBO basic and nested. With few exceptions, EIPU performs just as well if not better than the manually tuned algorithms.}
%     \label{fig:algorithms_performance}
% \end{figure}

\section{Conclusion}\label{sec:conclusion}
In this paper, we have looked at a generalizable version of cost-aware problems in light of the limitations encountered in real-world experiments requiring physical resources. 
We have analysed the behavior of multiple algorithms when a setup switch is associated with a cost and shown that 
the choice of switching throughout an optimisation process is non-trivial.
This number is primarily influenced by the dimensionsionality of the problem in relation to the number of costly dimensions, and the cost associated with each switch. 

We have adapted a number of algorithms that can be used to solve the above class of problems. The parameter-free EIPU algorithm proposed as one of the solution methods was shown to yield comparable results with other finely tuned algorithms used in process-constrained BO problems. The practical impact of these algorithms can help reduce the cost of optimization processes in several fields, from automotive to biopharmaceutical settings. 

A potential extension of the work presented in this paper comes from the inherent characteristic of BO being only interested in finding the next best point to evaluate. 
In our case, before evaluating the next point we want to know both how good that point is and its evaluation cost. 
In other words, analysing an evaluation point is done through an aggregate between cost and value. 
It thus becomes important for BO to understand the overall quality of a certain setup before committing to it. 
% It is assumed that once a switching cost is incurred, the ideal scenario will make use of multiple evaluations within the same setup before changing it. 
A potential solution worth pursuing is the analysis of the performance lookahead-based heuristics, which take into consideration how good a setup might be and make BO as a setting less greedy. 
% Likewise, future work can look at understanding the impact of the budget on the algorithm performance. 

\bibliographystyle{splncs04nat}
\bibliography{main}

\end{document}